\documentclass[pdftex,a4paper,twocolumn,twoside,10pt]{ranlp}
\usepackage[greek,russian,british]{babel}

\usepackage{natbib}               % natbib front-end
\usepackage{xspace}               % \xspace command

\usepackage[breaklinks,colorlinks,linkcolor=black,citecolor=black,pagecolor=black,urlcolor=black,anchorcolor=black,plainpages=false,pdfpagelabels]{hyperref}

\hypersetup{%
  pdfauthor={Stasinos Konstantopoulos},
  pdftitle={What's in a name?},
  pdfkeywords={language identification, language guessing, morphophonology, graphotactics},
  pdfsubject={This paper describes experiments on identifying the
    language of a single name in isolation or in a document written in
    a different language. A new corpus has been compiled and made
    available, matching names against languages. This corpus is
    used in a series of experiments measuring the performance of
    general language models and names-only language models on the
    language identification task. Conclusions are drawn from the
    comparison between using general language models and names-only
    language models and between identifying the language of isolated
    names and the language of very short document fragments. Future
    research directions are outlined.}
}

\usepackage[usenames]{color}

\usepackage{url}

\newcommand{\term}[1]{\emph{#1\/}}
\newcommand{\quotes}[1]{`#1'}
\newcommand{\ortho}[1]{[#1]}
\newcommand{\algo}[1]{\textsc{#1}}

\newcommand{\tmc}{Transfermarkt corpus\xspace} 
\newcommand{\ngram}{$n$-gram\xspace} 
\newcommand{\ngrams}{$n$-grams\xspace}

\begin{document}

\title{\textbf{What's in a name?}}

\author{
Stasinos Konstantopoulos     \\
IIT, NCSR \quotes{Demokritos} \\
Ag. Paraskevi 15310, Athens, Greece \\
Tel: +30 21 06503162 \\
Fax: +30 21 06532175 \\
\email{konstant}{iit.demokritos.gr}
}

\date{} % ranlp.cls instructions: keep this empty

\maketitle

\begin{abstract}
  This paper describes experiments on identifying the language of a
  single name in isolation or in a document written in a different
  language. A new corpus has been compiled and made available,
  matching names against languages. This corpus is used in a series of
  experiments measuring the performance of general language models and
  names-only language models on the language identification task.
  Conclusions are drawn from the comparison between using general
  language models and names-only language models and between
  identifying the language of isolated names and the language of very
  short document fragments. Future research directions are outlined.
\end{abstract}


\section{Introduction}
\label{sec:intro}

Language identification is performed on different levels, from the
acoustic and prosodic to the phonotactic or graphotactic, and has
found various application in speech synthesis, information extraction
and data mining.

Leaving aside language identification at the acoustic and prosodic
level, we shall concentrate on identifying the language of a string of
phonemes or graphemes. In fact, all of the methods and experiments
presented here operate on graphemes, but there is no underlying
assumption that forces this choice, only data availability.

We further concentrate on identifying the language of a single name,
even when it is in isolation or in a document written in a different
language. This is particularly interesting to named-entity
recognition, especially if the methodology supports spotting names
transliterated into different orthography systems, e.g. spotting
English-language named-entities in Chinese newspapers.

The intuition and basic hypothesis that the work presented here tests,
is that names are more \quotes{characteristic} of their language than
general words, and that a single name might have enough clues to
confidently identify its language, where a general word of the same
length wouldn't.

The paper is structured as follows: first an overview of the
literature in language identification is provided, both in the
framework of text categorization and for identifying the language of a
single named entity in isolation (Sect.~\ref{sec:bg}). Then, in
Sect.~\ref{sec:tmc} the corpus used is presented, as well as the
methodology for compiling it. Finally, the experimental setup and
results are described (Sect.~\ref{sec:experiments}) and conclusions
and future research directions are offered (Sect.~\ref{sec:conc}).

\section{Language Identification}
\label{sec:bg}

\subsection{Text Categorization}
\label{sec:bg:text}

Guessing the language of a document falls under the larger area of
\term{text categorization}, which aims at classifying a document as
belonging to one (or more) out of certain, predefined categories or
subject codes. Document language is one of the possible dimensions of
categorization, interesting for various document organization, data
mining, and information extraction tasks.

\citet{cavnar-etal:1994} report experiments on language categorization
using a simple \ngram frequency algorithm. The language models consist
of frequency counts of \ngrams (up to $5$-grams) for various languages.
To classify a document, the frequency counts of \ngrams in the
document are calculated, and their distribution compared against the
distribution of \ngrams in the language models. The model with the
smallest distance from the distribution of the document, is assumed to
be the language of the document.

This algorithm was tested on Usenet postings from the
\texttt{soc.culture} newsgroup hierarchy. An eight-language corpus was
generated semi-automatically: a first pass operated under the
assumption that the postings are in the language of the country or
region under discussion in each newsgroup, and at a second pass
discrepancies between the newsgroup's default language and the
system's prediction were manually resolved.

With the 400 most frequent \ngrams retained in the models, and
postings of at least 300 bytes of length, the system classified the
test set almost perfectly, achieving and accuracy of 99.8\%.
The authors also report an accuracy of 99.3\% for postings
that are under 300 bytes, without providing any further details of how
accuracy drops with shorter test documents.

\citeauthor{cavnar-etal:1994}'s algorithm has seen various
implementations and use cases, most notably the 
\algo{TextCat}\footnote{See \url{http://www.let.rug.nl/\~vannoord/TextCat/}}
implementation used in the
\algo{SpamAssassin}\footnote{See \url{http://spamassassin.apache.org/}} spam
filter. The \algo{TextCat} distribution includes language models for 69
languages and about 9~kbytes of text in each language.

\subsection{Language Identification in Isolation}
\label{sec:bg:names}

Language identification is very accurate even for texts as small as
two or three hundred characters, but even so that is a long way from
identifying the language of origin of single name, when seen in
isolation.

Efforts at language identification for proper names originate
in speech synthesis \citep{spiegel:1985,vitale:1991,llitjos-etal:2001},
with language identification used to adjust grapheme-to-phoneme rules. 
The typical approach is to improve an English-language speech
synthesizer by training \ngram classifiers and using different
pronunciation models for foreign names, depending on each name's
origin.

\citeauthor{llitjos-etal:2001}, in particular, note that language
identification of isolated names is a
difficult task, as they tried to manually tag 516 names and found that
they could confidently tag only 43\% of the data.
For their speech synthesis experiment they used a simplification of the
Cavnar and Trenkle algorithm which only counted $3$-grams. They trained
language models on general text (ranging from 255 thousand to 11
million words), and provided the classification results as features
for the grapheme-to-phoneme models. Unfortunately they do not report
results for the language identification part of their experiments, but
they do make the following observation:
\begin{quote}
  Ideally, we should have trained our LLMs [letter language models] on
  just names, instead of text corpora, since that is the distribution
  of our training data. However, some experiments where we had LLM
  trained on both text and just proper names for German, French and
  Spanish have shown that the probability of the two LLM were very
  close, and it never happened that the LLM trained on text performed
  worse than the LLM trained on proper names.
\end{quote}

Another field of application of the same general methodology is
automatic transliteration of named-entities for the purposes of
machine translation \citep{huang:2005}, except that here language
identification adjusts transliteration models instead of
grapheme-to-phoneme ones. In Huang's experiment languages were grouped
together in clusters, where clustering was trained on the effect that
it had on the transliteration models. The resulting clusters
roughly corresponded to familiar language groupings
(Chinese, Romance, English-and-Dutch, Nordic).
Again, language identification models are
reported to improve accuracy, but no results are provided for the
language identification sub-task per se.

Finally, language identification is pertinent to
information-extraction tasks such as named-entity recognition.
In this context it is important to be able to identify the original
language of a named entity in order to be able to recognize
transliterated named entities. \citet{virga-etal:2003} report identifying
references to English-language named entities in Chinese text.
Their approach is to train a tri-gram model on Chinese transliterations of
English names and use it to pick out English-language named-entities.
Knowing that a string is an English word, the original orthography can
be more accurately guessed.

\section{The Transfermarkt Corpus}
\label{sec:tmc}

In order to test the hypothesis stated in the introduction, a corpus
of person's names matched with the language of each name was created.
The corpus is based on the Transfermarkt
web-site,\footnote{See \url{http://www.transfermarkt.de/} The corpus
  was compiled from a web crawl through the site performed on 7-6-2007.}
which features various information about 22966 football players,
including---most crucially for our purposes---their nationality.
The site has complete player information about several German leagues
and the top league from 21 other countries. After discarding
mixed-language nationalities (e.g. Belgian and Swiss) and sparse
nationalities, and after combining some nationalities under a single
language (U.K. \& Ireland; Serbia, Croatia \& Bosnia;
and Czech Republic and Slovakia\footnote{A preliminary experiment has
  shown Czeck and Slovak \emph{names} to be practically
  indistinguishable, despite the substantial differences between the
  Czeck and Slovak \emph{languages}.})
there are $13$ languages left with a
reasonable\footnote{The cut-off point was set at $556$ examples, which
  leaves $500$ training examples at 10-fold cross-validation. Going
  for a minimum of $1000$ training examples would leave us with too
  few languages, including only one pair of closely related languages
  (Italian and French), rendering the experiments considerably less
  interesting. Also note the Greek would have been included if it were
  for corpus size alone, but was dropped since, retaining the original
  orthography, a single character is always enough to
  accurately identify it, making any comparison pointless.}
number of names each, listed in Table~\ref{tab:corpus-stats}.

\begin{table}
\begin{center}
\begin{tabular}{|l|rc|}
\hline
Language     & Names & Avg. Len. \\
\hline
German       &  3153 & 15.1 \\
English      &  1660 & 13.6 \\
Serbocroatian&  1474 & 14.3 \\
Italian      &  1151 & 16.2 \\
French       &  1141 & 15.8 \\
Polish       &  1057 & 16.0 \\
Spanish      &  1031 & 14.0 \\
Danish       &   817 & 15.7 \\
Dutch        &   809 & 15.1 \\
Swedish      &   746 & 15.7 \\
Czechoslovak &   653 & 13.6 \\
Norwegian    &   622 & 16.2 \\
Portuguese   &   600 & 11.1 \\
\hline
Total        & 14914 & 14.8 \\
\hline
\end{tabular}
\end{center}
\caption{Corpus size statistics for full names.}
\label{tab:corpus-stats}
\end{table}

A second dataset was created were only last names were considered. In
the Transfermarkt web-site full names are provided without any
indication of how they should be split into their first name/last name
components, so, for names with more than two parts, the last part was
assumed to be the last name. This assumption makes the task slightly
more difficult, since it removes language-specific surname prefixes
like \ortho{van} and \ortho{della}, but is accurate in most cases
since middle names are far more widespread than surname prefixes or
double surnames. This process has resulted in several duplicates, as
some people shared the same surname, and the subsequent shrinking of
the dataset. The corpus size statistics for last names are given
in Table~\ref{tab:lastname-corpus-stats}. Doing the same for first
names did not yield a usable-sized dataset, since repetitions
are considerably more frequent with first names. At the two extremes,
German was left with $609$ examples and Czechoslovak with $99$.

\begin{table}
\begin{center}
\begin{tabular}{|l|rc|}
\hline
Language     & Names & Avg. Len. \\
\hline
German       &  2608 & 7.8 \\
Serbocroatian&  1160 & 7.8 \\
English      &  1132 & 7.5 \\
French       &  1067 & 7.7 \\
Italian      &  1042 & 8.2 \\
Polish       &   944 & 8.6 \\
Spanish      &   824 & 7.4 \\
Dutch        &   746 & 7.5 \\
Czechoslovak &   579 & 7.2 \\
Swedish      &   542 & 8.6 \\
Danish       &   501 & 8.2 \\
Norwegian    &   488 & 7.8 \\
Portuguese   &   418 & 6.3 \\
\hline
Total        & 12051 & 7.8 \\
\hline
\end{tabular}
\end{center}
\caption{Corpus size statistics for last names.}
\label{tab:lastname-corpus-stats}
\end{table}

In the web-site, names originally spelled in Cyrillic or Greek are
transliterated, but the various diacritics used in Latin-based
languages are retained. All such diacritics were dropped from the
corpus, since some are sufficient to considerably narrow the problem
down or even identify a single language (e.g. Czech \ortho{\v r}).

The \tmc, both with diacritics and the plain Latin version, can be 
downloaded from \url{http://www.iit.demokritos.gr/~konstant/dload/tmc.tgz}
The archive includes the tools developed to compile and manipulate the
corpus and conduct the experiments described below.

\section{Experiments and Results}
\label{sec:experiments}

\subsection{Categorization of Short Strings}

As already mentioned, previous work has reported considerable
performance increase in transliteration and grapheme-to-phoneme
conversion tasks when language identification is applied before the
main algorithm, even for identifying the language of a single name.
Raw language guessing results, however, have not been reported except
by \citeauthor{cavnar-etal:1994}, but they only differentiate between
strings that are shorter or longer than 300 bytes.

In order to establish a basis for comparison, it was necessary to
conduct some experiments on language identification from strings of
length comparable to the average length of 15 characters observed in
the \tmc. The \algo{TextCat} implementation of the 
\citeauthor{cavnar-etal:1994} algorithm was used.
The test sets were built
by splitting the short texts included in the \algo{TextCat}
distribution. Various text fragment lengths were tried, ranging
between $5$ and $25$ characters, not necessarily aligned with word
boundaries.

Two sets of tests were run, one with full diacritics and one on plain
text. For the former tests, the models that come with the
\algo{TextCat} distribution were used. For the latter, the test sets
were stripped of diacritics and plain-text models were built from
100~kbytes fragments of the JRC-Acquis Multilingual Parallel Corpus
\citep{steinberger-etal:2006}. The JRC-Acquis corpus does not include
Norwegian and Serbocroatian from the languages of
Table~\ref{tab:corpus-stats}, so the models from \algo{TextCat} were
used for these languages. Serbian, Croatian, and Bosnian have
different models in \algo{TextCat}, so all three were included and the
results adjusted accordingly to consider any of these three answers as
correct. These models (and the test texts) are plain Latin
transliterations of the original orthographies, so they could be
immediately included.

The Norwegian model, on the other hand, is encoded with full
diacritics and had to be simplified. This has had a slight negative
effect on Norwegian results, since the frequencies in the new model
are not guaranteed to be accurate.
To illustrate why this is the case, consider the
$2$-gram \ortho{p\aa} which is present among the $400$ most
frequent \ngrams retained in the model, but \ortho{pa} is not.
In the model generated by merging \ortho{\aa} into \ortho{a},
\ortho{pa} is under-represented. In a more extreme case, an \ngram
might be altogether missing. Imagine, for example, a situation where
neither \ortho{pa} nor \ortho{p\aa} were frequent enough to be in the
original model, but their combined frequency would have been enough.

\begin{table}
\begin{center}
\begin{tabular}{|r|cc|}
\hline
String & Plain & Full \\
Length & Latin & Diacritics \\
\hline
5      & $32\%$      & $41\%$ \\
10     & $49\%$      & $62\%$ \\
15     & $60\%$      & $73\%$ \\
20     & $65\%$      & $78\%$ \\
25     & $69\%$      & $79\%$ \\
\hline
\end{tabular}
\end{center}
\caption{$F_{\beta=1}$ score of language identification of short
  strings from generic models.} 
\label{tab:textcat}
\end{table}

The predictive performance, averaged over $13$ languages, of the plain
Latin and the full-diacritics models is given in
Table~\ref{tab:textcat}. It should be noted that the Serbocroatian
model and test text in \algo{TextCat} is only available without
diacritics, so the same Serbocroatian text and models were employed in
both experiment series. Performance is given as the balanced $F$-score
over precision $P$ and recall $R$:
\begin{equation}
\nonumber
F_{\beta=1} = \frac{2 \cdot P \cdot R}{P + R}
\end{equation}

The most immediate conclusions from these results is that at this size
range language identification has not converged yet at the numbers
reported at the $300$-character area, and that diacritics have a very
high discriminative power. Both conclusions were intuitively
anticipated, and these results are mostly reported in order to
co-relate them with the main results on language identification of names.

\subsection{Categorization of Names}

Having established the performance of general-purpose \algo{TextCat}
models on general text, the next step is to measure their performance
over the \tmc and compare it against the performance of models
specifically trained on names.

\begin{table}
\begin{center}
\begin{tabular}{|lr|ccc|}
\hline
Language     & Avg  & Recall & Prec. & $F_{\beta=1}$ \\
             & Len  &        &       & \\
\hline
Italian      & 16.2 & 37\% & 62\% & 46\% \\
Norwegian    & 16.2 & 13\% & 16\% & 14\% \\
Polish       & 16.0 & 65\% & 67\% & 66\% \\
French       & 15.8 & 11\% & 36\% & 17\% \\
Swedish      & 15.7 &  3\% &  5\% &  4\% \\
Danish       & 15.7 & 14\% & 17\% & 15\% \\
German       & 15.1 & 41\% & 39\% & 40\% \\
Dutch        & 15.1 & 22\% & 46\% & 30\% \\
Serbocroatian& 14.3 & 38\% & 57\% & 46\% \\
Spanish      & 14.0 & 18\% & 23\% & 20\% \\
English      & 13.6 &  7\% & 22\% & 11\% \\
Czechoslovak & 13.6 &  5\% & 20\% &  8\% \\
Portuguese   & 11.1 & 29\% & 30\% & 29\% \\
\hline
Average      & 14.8 & 23\% & 34\% & 27\% \\
\hline
\end{tabular}
\end{center}
\caption{Performance of general language identification models over
  full names.} 
\label{tab:names-generic}
\end{table}
\begin{table}
\begin{center}
\begin{tabular}{|lr|ccc|}
\hline
Language     & Avg  & Recall & Prec. & $F_{\beta=1}$ \\
             & Len  &        &       & \\
\hline
Italian      & 16.2 & 66\% & 69\% & 67\% \\
Norwegian    & 16.2 & 41\% & 36\% & 38\% \\
Polish       & 16.0 & 80\% & 88\% & 84\% \\
French       & 15.8 & 47\% & 50\% & 48\% \\
Swedish      & 15.7 & 51\% & 47\% & 49\% \\
Danish       & 15.7 & 49\% & 58\% & 53\% \\
German       & 15.1 & 36\% & 71\% & 48\% \\
Dutch        & 15.1 & 44\% & 34\% & 38\% \\
Serbocroatian& 14.3 & 80\% & 79\% & 79\% \\
Spanish      & 14.0 & 45\% & 55\% & 50\% \\
English      & 13.6 & 58\% & 70\% & 63\% \\
Czechoslovak & 13.6 & 70\% & 55\% & 62\% \\
Portuguese   & 11.1 & 52\% & 60\% & 56\% \\
\hline
Average      & 14.8 & 27\% & 55\% & 60 \%\\
\hline
\end{tabular}
\end{center}
\caption{Performance of 10-fold cross-validated full name models} 
\label{tab:fullnames-10fold}
\end{table}
\begin{table}
\begin{center}
\begin{tabular}{|lr|ccc|}
\hline
Language     & Avg  & Recall & Prec. & $F_{\beta=1}$ \\
             & Len  &        &       & \\
\hline
Italian      & 8.1 & 61\% & 59\% & 60\% \\
Norwegian    & 7.8 & 30\% & 31\% & 30\% \\
Polish       & 8.6 & 61\% & 75\% & 67\% \\
French       & 7.7 & 39\% & 36\% & 37\% \\
Swedish      & 8.6 & 51\% & 61\% & 56\% \\
Danish       & 8.2 & 42\% & 60\% & 49\% \\
German       & 7.8 & 44\% & 73\% & 55\% \\
Dutch        & 7.5 & 38\% & 35\% & 36\% \\
Serbocroatian& 7.8 & 74\% & 79\% & 76\% \\
Spanish      & 7.4 & 38\% & 41\% & 39\% \\
English      & 7.5 & 51\% & 66\% & 58\% \\
Czechoslovak & 7.2 & 56\% & 36\% & 44\% \\
Portuguese   & 6.3 & 40\% & 41\% & 40\% \\
\hline
Average      & 7.8 & 48\% & 53\% & 50\% \\ 
\hline
\end{tabular}
\end{center}
\caption{Performance of 10-fold cross-validated last name models} 
\label{tab:lastnames-10fold}
\end{table}

As a first step, the JRC-Acquis models were tested on the
Transfermarkt names, where they performed substantially worse than they
did on the \algo{TextCat} test texts (cf. Table~\ref{tab:names-generic}).
This is, to a large extent, due to the fact that the nationality of
the bearer of a name does not consistently reflect the name's origin,
resulting in considerable noise in the Transfermarkt dataset.

At a second step, the full name and last name datasets were used to
train and test the language models, using 10-fold cross-validation.
$N$-fold cross-validation is a methodology for evaluating a hypothesis
when there is not enough data to obtain both a training and a test
set, but the same data has to be used for both training and
validation, while at the same time guaranteeing the independence of
the training and the validation process. The original set is
partitioned into $N$ subsets, of which one is retained as testing data
and the remaining $N - 1$ are used as training data. Training and
testing is repeated $N$ times (the folds), with
each of the $N$ subsets used exactly once as testing data. The $N$
results from the folds are averaged to produce a single estimation.

The results of the full name and last name language models
(Tables~\ref{tab:fullnames-10fold}
and~\ref{tab:lastnames-10fold}, resp.) show a completely different
picture than the JRC-Acquis models,
where performance over plain-Latin general text is matched and at times
surpassed. Comparing full-name language identification against
language identification of general text we see that similar results
are obtained, despite the fact that the names dataset is a
considerably more noisy, as shown by the performance of the JRC models
over the Transfermarkt dataset. Last-name language identification
performs even better, almost touching the results of general text
language identification \emph{with} diacritics.

%\ranlpfig{0.45}{ranlp07-scores}{fig:scores}{Results}

\begin{figure*}
\begin{center}
  \mbox{\epsfxsize=0.65\textwidth\epsfbox{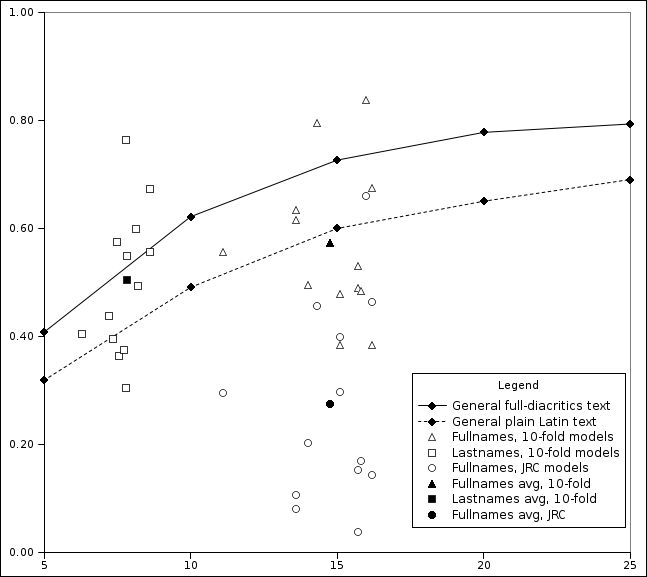}}
\end{center}
\caption{Graph plotting F-score of language identification against
  string length. The two lines plot language identification
  performance over general text of fixed length. The outlined circle,
  triangle, and square marks show F-scores per language, against
  average name length of the language. The filled circle, triangle,
  and square marks average these last results over all 13
  languages.}
\label{fig:results}
\end{figure*}

By comparing full name and last name results, we observe that,
although full names are on average almost twice as long as last names,
last names alone are enough to achieve an F-score of 50\%, versus 60\%
for full names. Intuitively, this is corroborated by the fact
(observed in Sec.~\ref{sec:tmc} above) that there are a lot more distinct
last names than there are first names, so that the former are more
\quotes{dense} in information that the latter.

Finally, comparing the results of the JRC models and the Transfermarkt
models of full names, tested over the Transfermarkt names, we
see that training models specific to names has a most profound effect
on performance, with the average F-score more than doubling (27\%
versus 60\%). This is in sharp contrast to the remark by
\citeauthor{llitjos-etal:2001} (cf. Sect.~\ref{sec:bg:names})
that they did not observe any performance increase when training
language models with datasets of names instead of general text.

Figure~\ref{fig:results} combines all of these results into a graph
were the, so to speak, relative \quotes{discriminative density}, of
names and general words can be seen. For full names, it is practically
identical to general words since full names have an average length of
almost 15 characters and can be predicted at pretty much the same rate
as 15-character-long general words. First names, on the other hand,
carry a lot more potential per character, as their average length is
just under 8 characters, but can be predicted as accurately as
general words of about 11 characters.

\section{Conclusions and Future Research}
\label{sec:conc}

The main conclusion drawn from the experiments exposed here is that
names offer themselves for more accurate language identification than
general words. This conclusion has been repeatedly hinted at in
previous work on grapheme-to-phoneme conversion and transliteration,
where a language identification pre-processing step resulted in
dramatic performance increase on the main task.

Furthermore, it has been demonstrated that language identification
models for names work better when trained on names, despite prior
reports to the contrary in the literature. It is interesting to note
that the performance reported here is on a par with the performance of
human annotators, who reported that they could only confidently
predict a person's nationality in 43\% of the data
(cf. Sect.~\ref{sec:bg:names}).

Finally, it has been confirmed that last names carry more information
per character than first names. This result which was expected, since
it is a lot more common for first names to appear in multiple
languages than it is for last names.

A few open questions remain, however. One is the nature of the
information that a name (or any word in general) carries that allows
humans and Machine Learning systems to predict its linguistic
background. The information used can be either morphological or
phonotactic/graphotactic. For example, characteristic suffixes or
prefixes like Konstant\emph{opoulos} or \emph{Mc}Leod are
morphological features of their respective languages, whereas, say, a
consonant cluster like Polish \ortho{krz}, only found in a single
(small set of) language(s), is a phonotactic phenomenon, convolved
with the orthographic conventions of the language(s) at hand.

An interesting line of research to pursue is devising and conducting
experiments that would analyse the contribution of each of these factors
to the observed increase in language identification performance when
dealing with names. There are various preliminary thoughts along these
lines, including using more explicit models, combined with using
representations that are more cross-linguistically uniform and
uninformed with respect to the spelling of the name in its original
language.

What is meant by more explicit models, is \emph{model representations}
where different types of features are used instead of a flat \ngram
model. So, for example, one could imagine that interesting conclusions
could be drawn by analysing a model that has access to prior
phonotactic models, grapheme-to-phoneme mapping models, and the
results of derivational morphology analysis. This, however, would be a
major undertaking requiring a long array of linguistic resources,
which can be very sparse for some of the languages discussed here.

The second idea mentioned above is that of a cross-linguistically
uniform and uninformed \emph{data representation}. In the experiments
presented here a small step in this direction was taken by dropping
all diacritics, so that there will be fewer chances for \quotes{easy
  guesses} based on characters only found in a single language. This
creates a performance mis-balance in favour of orthographies that
prefer grapheme clusters instead of overloaded characters, as, for
example, the distinction between \ortho{s} and \ortho{sh} is retained
in English and the distinction between \ortho{s} and \ortho{sz} is
retained in Polish, but the distinction between \ortho{s} and
\ortho{\v s} is lost in Czechoslovak and Serbocroatian.

One idea would be to represent names in an abstract phonological
representation, but that creates the additional problem of devising
such a representation which is neither too detailed nor too coarse and
accurately making all the grapheme-to-phoneme conversions necessary.
An attractive alternative would be based on the
assumption that transliteration to a completely different orthography
to a large extend removes clues that are based on orthographic
idiosyncrasies of the original language. So, for example, if a
resource similar to the \tmc would be created from a Greek or Russian
site, all instances of \ortho{sh}, \ortho{sz} and \ortho{\v s} would
have been spelled as \ortho{\foreignlanguage{greek}{sv}}
and \ortho{\foreignlanguage{russian}{\cyrsh}},
respectively.\footnote{Of course no transliteration is completely
  reversible and some information is lost. In the case of Greek the
  distinction between \ortho{s} and \ortho{sh-sz-\v s} is
  lost, and in Russian there is no English \ortho{th}, Spanish
  \ortho{z}, Greek \ortho{\foreignlanguage{greek}{j}}.}

Along a totally different line of research, other dimensions of
categorization could be explored. For example, depending on the
language at hand, separating male from female names can range from
trivial to impossible. Names can also provide indications about age
(as first names come and go into fashion), social and economical
status (compare \quotes{Paddy} and \quotes{Patrick III}), religious
and, in general, cultural background
(Ren\'e\ Antonius Maria Eijkelkamp and Abdelhali Chaiat are both Dutch
football players, but certain educated guesses can be made about their
cultural backgrounds from their names alone), etc.

The extend to which such categorization attempts will remain
morphophonologically interesting and will not deteriorate into
thesaurus look-ups or require full-blown pragmatic knowledge cannot be
predicted, but might be worth investigating.

\section{Acknowledgements}

The work described here was partially supported by MedIEQ,
an EU-funded project.\footnote{DG-SANGO Programme \quotes{Public Health},
  Action~1.5 \quotes{eHealth}. See also \url{http://www.medieq.org}}
MedIEQ investigates the automation of the quality labelling process in
medical web sites, by verifying their content using
machine-readable quality criteria.

\bibliographystyle{abbrvnat}

\bibliography{cs,ling}

\begin{thebibliography}{7}
\providecommand{\natexlab}[1]{#1}
\providecommand{\url}[1]{\texttt{#1}}
\expandafter\ifx\csname urlstyle\endcsname\relax
  \providecommand{\doi}[1]{doi: #1}\else
  \providecommand{\doi}{doi: \begingroup \urlstyle{rm}\Url}\fi

\bibitem[Cavnar and Trenkle(1994)]{cavnar-etal:1994}
W.~B. Cavnar and J.~M. Trenkle.
\newblock {N}-gram-based text categorization.
\newblock In \emph{Proceedings of Third Annual Symposium on Document Analysis
  and Information Retrieval, Las Vegas, 11--13 April 1994}, pages 161--175,
  1994.

\bibitem[Font~Llitj{\'o}s and Black(2001)]{llitjos-etal:2001}
A.~Font~Llitj{\'o}s and A.~W. Black.
\newblock Knowledge of language origin improves pronunciation accuracy of
  proper names.
\newblock In \emph{Proc. of Eurospeech 2001, Aalborg, Denmark}, 2001.
\newblock URL
  \url{http://sidecar.sp.cs.cmu.edu/ari/papers/eurospeech01/Eurospeech01.pdf}.

\bibitem[Huang(2005)]{huang:2005}
F.~Huang.
\newblock Cluster-specific named entity transliteration.
\newblock In \emph{HLT '05: Proceedings of the conference on Human Language
  Technology and Empirical Methods in Natural Language Processing, Vancouver,
  British Columbia, Canada}, pages 435--442, Morristown, NJ, USA, 2005.
  Association for Computational Linguistics.

\bibitem[Spiegel(1985)]{spiegel:1985}
M.~F. Spiegel.
\newblock Pronouncing surnames automatically.
\newblock In \emph{Proc. Conf. of the American Voice Input/Output Society},
  pages 109--132, 1985.

\bibitem[Steinberger et~al.(2006)Steinberger, Pouliquen, Widiger, Ignat,
  Erjavec, Tufi\c{s}, and Varga]{steinberger-etal:2006}
R.~Steinberger, B.~Pouliquen, A.~Widiger, C.~Ignat, T.~Erjavec, D.~Tufi\c{s},
  and D.~Varga.
\newblock The {JRC-Acquis}: A multilingual aligned parallel corpus with 20+
  languages.
\newblock In \emph{Proc. of the 5th International Conference on Language
  Resources and Evaluation (LREC'2006), Genoa, 24--26 May 2006}, 2006.

\bibitem[Virga and Khudanpur(2003)]{virga-etal:2003}
P.~Virga and S.~Khudanpur.
\newblock Transliteration of proper names in cross-lingual information
  retrieval.
\newblock In M.~Broman~Olsen, editor, \emph{Proceedings of the ACL 2003
  Workshop on Multilingual and Mixed-language Named Entity Recognition}, pages
  57--64, 2003.
\newblock URL \url{http://www.aclweb.org/anthology/W03-1508.pdf}.

\bibitem[Vitale(1991)]{vitale:1991}
T.~Vitale.
\newblock An algorithm for high accuracy name pronunciation by parametric
  speech synthesizer.
\newblock \emph{Computational Linguistics}, 17\penalty0 (3):\penalty0 257--276,
  1991.

\end{thebibliography}

\end{document}